\documentclass[runningheads]{llncs}
\usepackage[T1]{fontenc}
\usepackage{makecell}
\usepackage{amsmath}
\usepackage{eqnarray}
\usepackage{amssymb}							
\usepackage{lmodern}							
\usepackage{xcolor}
\usepackage{multirow}
\usepackage{booktabs}
\usepackage{graphicx}
\usepackage{threeparttable}
\usepackage{graphicx}
\begin{document}
\title{A Methodological Report on Anomaly Detection on Dynamic Knowledge Graphs}
\titlerunning{Anomaly Detection on Dynamic Knowledge Graphs}
%
\author{Xiaohua Lu \thanks{corresponding author: xh.lu0706@gmail.com}\and Leshanshui Yang \inst{1,}
}
\authorrunning{X. Lu et al.}
%
\institute{University of Rouen Normandy, LITIS UR 4108, 76000 Rouen, France} 
\maketitle              
\begin{abstract}
In this paper, we explore different approaches to anomaly detection on dynamic knowledge graphs, specifically in a Micro-services environment for Kubernetes applications. Our approach explores three dynamic knowledge graph representations: sequential data, hierarchical data and inter-service dependency data, with each representation incorporating increasingly complex structural information of dynamic knowledge graph. Different machine learning and deep learning models are tested on these representations. 
We empirically analyse their performance and propose an approach based on ensemble learning of these models. Our approach significantly outperforms the baseline on the ISWC 2024 Dynamic Knowledge Graph Anomaly Detection dataset, providing a robust solution for anomaly detection in dynamic complex data.
\end{abstract}

\section{Introduction}

With the development of graph machine learning and geometric deep learning \cite{bronstein2017geometric}, research on encoding and prediction has been applied to more domains than ever before. Graphs can simultaneously represent entities through nodes\footnote{Also referred to as "vertices".} and relationships between entities through edges. As a result, graph representations are widely used in domains as diverse as transportation \cite{yu2017spatio}, chemistry \cite{gilmer2017neural}, and social networks \cite{TGN}. Two areas of relative difficulty in graph learning are heterogeneous graphs and dynamic graphs. Heterogeneous graphs allow for different types of nodes and/or edges in the graph \cite{shi2022heterogeneous}. They can therefore better model complex systems with different types of entities, such as a distributed information system with services, pods and nodes. Dynamic graphs allow the structure and/or attributes of a graph to change over time \cite{yang2024dynamicsurvey}, allowing better modelling of data with both temporal and structural information.

Dynamic Knowledge Graphs (DKG), as an intersection of heterogeneous and dynamic graphs, have attracted increasing attention due to their ability to represent the evolving nature of knowledge over time. By exploiting the ability to capture topological and attribute transformations, DKGs provide valuable insights into the evolution of knowledge in diverse domains such as Social Media \cite{dorpinghaus2022social}, the Internet of Things (IoT) \cite{IOTsteenwinckel2021flags,IOTsteenwinckel2022data}, and E-commerce.

Kubernetes-based Micro-services Architecture (MSA) platforms consist of interconnected entities, such as compute nodes, deployable units (pods), and services, which work together to build applications like e-commerce platforms. Since these entities are different and they also depend on each other through different relationships, the best way to represent the data information and the structural information is to construct a Knowledge Graph (KG) \cite{kg}. Moreover, the information and dependencies among entities can change over time in MSA platforms. Therefore, the use of a Dynamic Knowledge Graph (DKG) \cite{dkg} that records a complete snapshot of DKG at each time step is the best way to register and track these changes.

 In order to collect and store this DKG dataset, the organisers used the TTL format \footnote{https://en.wikipedia.org/wiki/Turtle\_(syntax)}, a textual syntax for the Resource Description Framework (RDF) that allows KG information to be represented in a compact textual form, such as a triple consisting of a subject, a predicate and an object \footnote{https://www.w3.org/TR/rdf12-turtle/\#language-features}. Using the Turtle language, the entities and relations of a snapshot of each time step are stored in ternary relations. Predictions based on this structured data is crucial for the operation and optimisation of MSA. This ternary structure representation is similar to the syntax of natural language, which helps to improve the efficiency of querying and parsing, thus improving the effectiveness and robustness of the anomaly detection model.


Although DKGs are well suited for modelling different nodes, variant relationships, and temporal information in Kubernetes-based MSA platforms, their analysing and learning approaches are not well explored yet. In addition, several challenges are specific to the anomaly detection task of Kubernetes-based MSA platforms: 1. Anomalous events are usually very rare, resulting in an imbalanced data distribution, which prevents the model's ability to learn and generalise anomalous features effectively. 2. Traditional RNNs and homogeneous GNNs are insufficient for accurate anomaly detection due to the need to capture temporal and structural information in Kubernetes environments, coupled with the presence of heterogeneous nodes and edges. To address these challenges, we need to develop specialised approaches capable of handling the dynamic and complex structures of Kubernetes-based DKGs. 


This paper introduces a novel anomaly detection framework for dynamic knowledge graphs, leveraging both graph structure and temporal dependencies. Our method integrates neighbourhood feature extraction with advanced machine learning techniques, including Support Vector Machine (SVM) \cite{SVM}, eXtreme Gradient Boosting (XGB) \cite{XGB}, and Self-Attention mechanism (SA) \cite{vaswani2017attentionneed} to enhance the model's ability to capture complex patterns of anomalies.
Our approach achieves significant performance improvements over the baseline in the Anomaly Detection on Dynamic Knowledge Graphs challenge (ADDKG)\footnote{https://eval.ai/web/challenges/challenge-page/2267/overview} at ISWC 2024, providing a robust solution for anomaly detection in dynamic and complex environments.

The structure of the paper is as follows. Section \ref{sec:related_work} gives the problem formulation and describes related work. Section \ref{sec:methodology} presents the methodology of the paper. Section \ref{sec:experiment} describes the experimental part of the work and discusses the results obtained. Section \ref{sec:conclusion} concludes our contribution to the ADDKG challenge and discusses future works.

\section{Related Work} \label{sec:related_work}

Recent studies have shown that the dynamic knowledge graph representation is expressive for anomaly detection in temporally complex systems. In this section, we review the related work in this area. In subsection \ref{secsub:DKG}, we present the definition and notations of the Dynamic Knowledge Graph. Then, in subsection \ref{secsub:AD}, we present common approaches to anomaly detection.

\subsection{Dynamic Knowledge Graphs} \label{secsub:DKG}
In static graphs, common representations include edge lists and adjacency matrices. Among them, the edge list is widely used in software development as it requires less memory. An edge list is a collection of edges. Each edge is represented as a pair of nodes.

In the realm of dynamic graphs, existing research categorises them into Continuous-Time Dynamic Graphs (CTDG) and Discrete-Time Dynamic Graphs (DTDG) \cite{yang2024dynamicsurvey}. CTDGs represent dynamic graphs using timestamped events, such as $(u, v, t)$, where $u$ and $v$ are nodes and $t$ is the timestamp of the interaction. In contrast, DTDGs represent dynamic graphs as a sequence of static graph snapshots $\{G_1, G_2, ..., G_T\}$.

Combined with the dynamic graph representations, 
the Dynamic Knowledge Graph (DKG) is modelled as a heterogeneous dynamic graph, represented by timestamped edge lists. 
Among the various formats for representing DKGs, Turtle format (TTL) is used for this competition. The TTL format allows a snapshot of DKGs to be modelled as a series of triples in the structure of (subject, predicate, object). Then a filename that records a timestamp is bound to this snapshot for registering the temporal information. 
An example of a TTL-based DKG at moment $t$ is given in Eq. \ref{eq:TTL}, where \(S\) is the ensemble of subject nodes, \(O\) is the ensemble of object nodes, \(P\) is the ensemble of predicates (or relationships), and \(T\) is the ensemble of timestamps.

\begin{eqnarray} \label{eq:TTL}
    \text{DKG}_t = \{(s, p, o) \mid s \in S, p \in P, o \in O\} 
    \nonumber \\ 
    \text{Timestamp } t \text{ is associated with DKG}_t , t\in T
\end{eqnarray}


In the problem studied in this paper, all anomalies are labelled at the graph level. Specifically, each anomaly in the DKG is labelled with a category $c$ occurring from $t_{start}$ to $t_{end}$.

\begin{table}[h]
\caption{Description of system faults and their types.}
\label{tab:anomaly_types}
\scriptsize
\centering
\begin{tabular}{|l|l|l|}
\hline
\textbf{Fault} & \textbf{Type} & \textbf{Description}   \\ \hline
Node CPU Hog & Static & Consumes the CPU resources of the targeted node. \\ \hline
\multirow{2}{*}{Node Drain} & Dynamic & Forces rescheduling of the resources running on \\
                            & & the targeted node. \\ \hline
\multirow{2}{*}{Node Memory Hog} & Static & Consumes the memory resources of the targeted \\
                                 & & node. \\ \hline
\multirow{2}{*}{Pod Autoscaler} & Dynamic & Changes the desired number of replicas for the \\
                                & & targeted workload. \\ \hline
\multirow{2}{*}{Pod CPU Hog} & Static & Consumes the CPU resources of the targeted \\
                             & & application container. \\ \hline
\multirow{2}{*}{Pod Delete} & Dynamic & Causes forced/graceful pod failure of a random \\ 
                            &  & replica of the targeted workload \\ \hline
\multirow{2}{*}{Pod Network Latency} & Static & Latency introduced in the network communication \\ 
                                    & &  between different pods. \\ \hline
\end{tabular}

\end{table}

\subsection{Anomaly Detection} \label{secsub:AD}

Anomaly detection is a crucial task in various fields, including network security, fraud detection, and financial systems \cite{bca,bco}. This task focuses on identifying anomalous patterns that deviate from the typical observation of the data \cite{adgnn_survey}.

Traditional anomaly detection methods typically rely on statistical techniques and distance-based measures. However, these methods often struggle with high-dimensional data and complex dependencies \cite{hodge2004survey}.
To overcome these problems, common machine learning approaches include SVM \cite{SVM}, XGB \cite{XGB}, and Isolation Forest (IF) \cite{IF}. SVM identifies the optimal hyperplane that maximises the distance between normal data points and the decision boundary, effectively distinguishing outliers. XGB is a gradient-boosting decision tree algorithm that is able to evaluate the importance of each feature and find those that are most effective in separating normal from abnormal data. IF is an unsupervised method that detects anomalies by constructing decision trees to isolate observations, with outliers requiring fewer splits to isolate due to their rarity and distinctiveness.

With the advancement of software algorithms and hardware, deep learning methods have been increasingly popular for anomaly detection due to their ability to extract complex features \cite{SurveyGADma}. Common architectures for time series include Recurrent Neural Networks (RNN) \cite{RNNJordanNetwork}, Temporal Convolutional Networks (TCN) \cite{oord2016wavenet}, and the Self-Attention mechanism (SA) \cite{vaswani2017attentionneed}. 
RNNs, including Long Short-Term Memory (LSTM) \cite{LSTM} and Gated Recurrent Unit (GRU) \cite{GRU}, capture temporal dependencies, addressing gradient issues in long sequences. TCNs handle long-term dependencies through causal convolutions, while SA captures long-range dependencies by examining relationships across sequences.



With the development and popularity of financial, social, traffic and biological networks, graph-based models have gained significant attention in anomaly detection tasks. Graph-based anomaly detection can be classified into four types \cite{adgnn_survey}: node anomalies, edge anomalies, subgraph anomalies and full graph anomalies. Early research \cite{dominant,cola,glocal,ocgtl} focused on static graph anomaly detection, mainly on node anomalies, with limited attention to edge and subgraph anomalies. However, real-world graph data tends to change over time, which introduces additional complexity. To address this problem, researchers have proposed dynamic graph solutions that aim to detect anomalies that occur as the graph changes. Some studies \cite{StrGNN,AddGraph,TADDY} have explored how to detect anomalies at edges or nodes in dynamic graphs. Moreover, networks not only evolve over time, but also present complex structures with heterogeneous nodes and edges. Knowledge Graphs (KGs) provide a powerful framework for modelling such heterogeneous graphs, making them suitable for anomaly detection tasks. Despite the potential of KGs, the application of anomaly detection to (dynamic) KGs is still limited, mainly due to the unstructured nature of KGs and the inherent challenges of dealing with their dynamic and heterogeneous nature.

The growth of cloud services and distributed systems has led to the widespread use of MSA in software and web applications, moving away from traditional monolithic architectures \cite{newman2021building}. MSA breaks down an application into independent services, each responsible for a specific function and able to communicate. This architecture enhances the scalability and flexibility of software, but as the number of services increases, it becomes increasingly difficult to manually identify anomalies. Therefore, an automated anomaly detection system is essential to maintain the system.

Most existing work in this area leverages three main data sources (logs \cite{yagoub2018equipment}, performance metrics \cite{sharma2013cloudpd}, and distributed traces \cite{pahl2018all}) and uses machine learning algorithms such as SVMs, K-Nearest Neighbours (KNN) \cite{knn}, Random Forests (RF) \cite{rf}, and Multi-Layer Perceptrons (MLPs)\cite{mlp} for anomaly detection. However, these methods often ignore the structural relationships between services. Modelling MSA as (dynamic) KGs can capture these inter-service dependencies and provide a more comprehensive framework for anomaly detection. However, the integration of DKGs with MSA for anomaly detection is still under-researched, mainly due to the challenges posed by the complexity and scalability of DKGs.

In addition to all these, the anomaly detection task itself presents challenges. Traditional machine learning methods struggle to capture the complex temporal and structural dependencies, while deep learning models require a large amount of labelled data and a balance between the amount of normal and abnormal data. However, in practice, the amount of anomalous data tends to be much less than the amount of normal data \cite{adgnn_survey}. Although techniques like negative sampling \cite{AddGraph,StrGNN,TADDY} can generate anomalous edges in dynamic graphs without attributed edges, this approach is infeasible for DKGs with attributed edges. Capturing both temporal and structural dependencies remains key to tackling this challenge.

To address these gaps, the ADDKG competition provides a dataset based on MSA and DKGs and encourages participants to develop innovative solutions for anomaly detection in this complex environment.


\section{Methodology} \label{sec:methodology}

This section introduces three representations of dynamic knowledge graphs: sequential data, hierarchical data, and inter-service dependency data, each adding increasing structural complexity. We then describe the machine learning and deep learning methods applied to these representations, followed by proposed ensemble learning  schemes. 

\subsection{Simplified Tree Structure} \label{subsec:tree}

\begin{figure}[!h]
    \centering
    \includegraphics[width=0.9\textwidth]{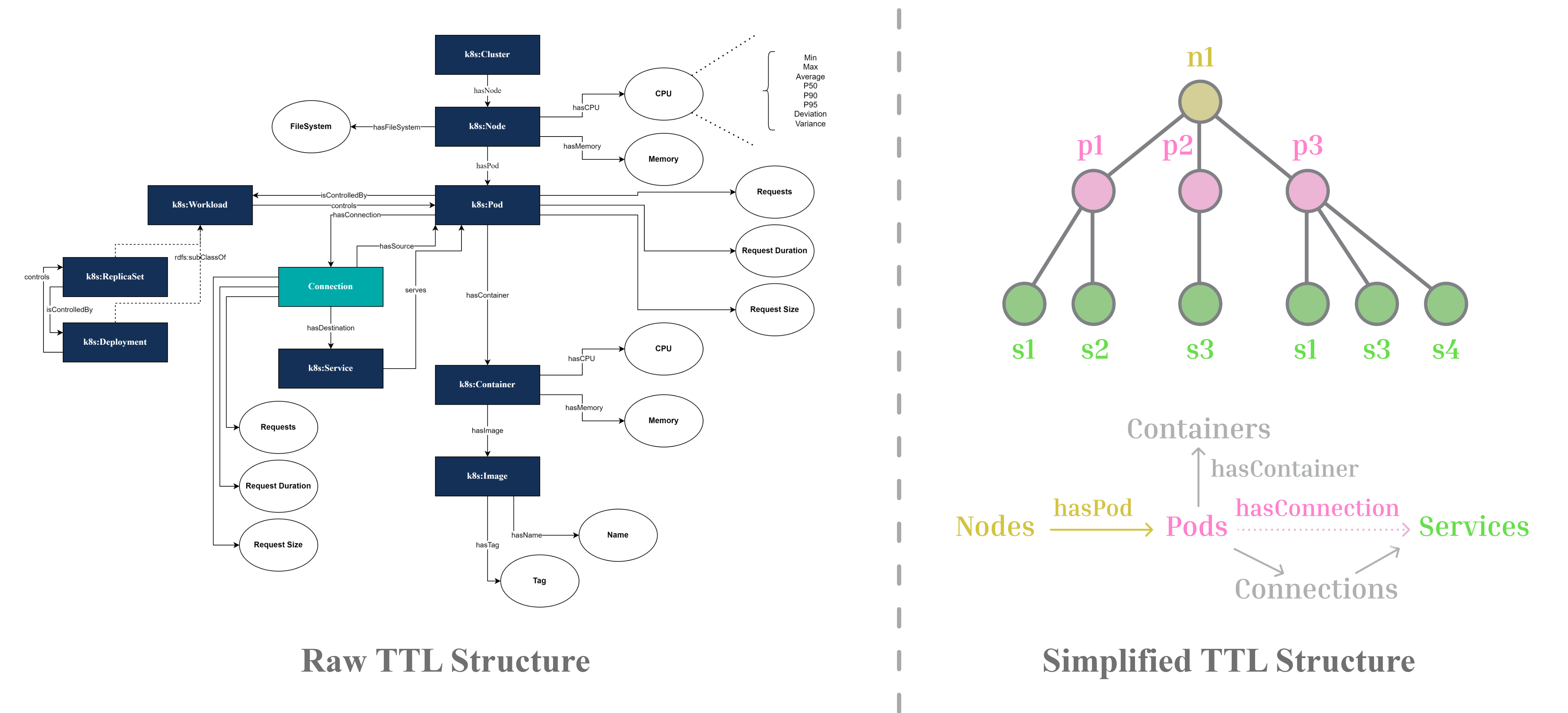}
    \caption{Left: The ontology overview of the relationships between Kubernetes concepts of the ADDKG dataset. Right: The simplified hierarchical structure of the ADDKG dataset. The container information is aggregated as the attributes of Pods. The Connections serve as attributes of the edges between Pods and Services.}
    \label{fig:input}
\end{figure}

The left part of Fig. \ref{fig:input} illustrates the ontology of the DKG utilised in the ADDKG dataset. The ontology comprises a total of ten distinct vertices (also known as functional units in ADDKG DKGs): Cluster, Node, Pod, Container, Image, Connection, Service, Workload, ReplicaSet and Deployment. These vertices are connected to each other with different relationships. Through a comprehensive analysis, three principal observations were identified:

\paragraph{\textbf{Primary Vertices.}} The ontology includes three primary vertices—Node, Container, and Connection. These are designated as primary vertices because only their properties record CPU performance metrics: minimum, maximum, mean, deviation, variance, 50th percentile, 90th percentile, and 95th percentile of CPU resource consumptions at each time step.

\paragraph{\textbf{Directional Relationships.}} All edges within the ontology are unidirectional. For example, relationships such as (Cluster, hasNode, Node) and (Node, hasPod, Pod) demonstrate this unidirectional nature. The only exception is the bidirectional edge pair "controls" and "is controlled by," which essentially represent the same relationship from opposite perspectives.

\paragraph{\textbf{Simplified Tree Structure.}} The DKG can be abstracted into a simplified tree structure, as depicted in the right part of Fig. \ref{fig:input}. This simplification consolidates information into three main vertices: Node, Pod, and Service, connected by the relationships (Node, hasPod, Pod) and (Pod, hasConnection, Service). 
According to the ontology, node attributes from the primary vertices (Container and Connection) can be effectively propagated to Pod and Service through the "hasContainer" and "hasConnection" relations.

The illustrated tree structure in Fig. \ref{fig:input}, comprising Node, Pod, and Service, serves as a reduced version of the tree structure, hence it can not represent the whole DKG. We will use this example to facilitate the later explanation of the three data representation methods used in our study. In the example, yellow vertices represent Nodes, pink vertices represent Pods, and green vertices represent Services. Each vertex type can have varying numbers of instances; for instance, this example includes one Node (n1), three Pods (p1, p2, p3), and four Services (s1, s2, s3, s4).


\subsection{Sequential Data Representation} \label{subsec:D1}

In this phase, the representation of the DKGs can be constructed by concatenating the time-evolving attributes of the vertices in the simplified tree structure, specifically from vertices Node, Pod and Service. 

\begin{figure}[!h]
    \centering
    \includegraphics[width=0.65\textwidth]{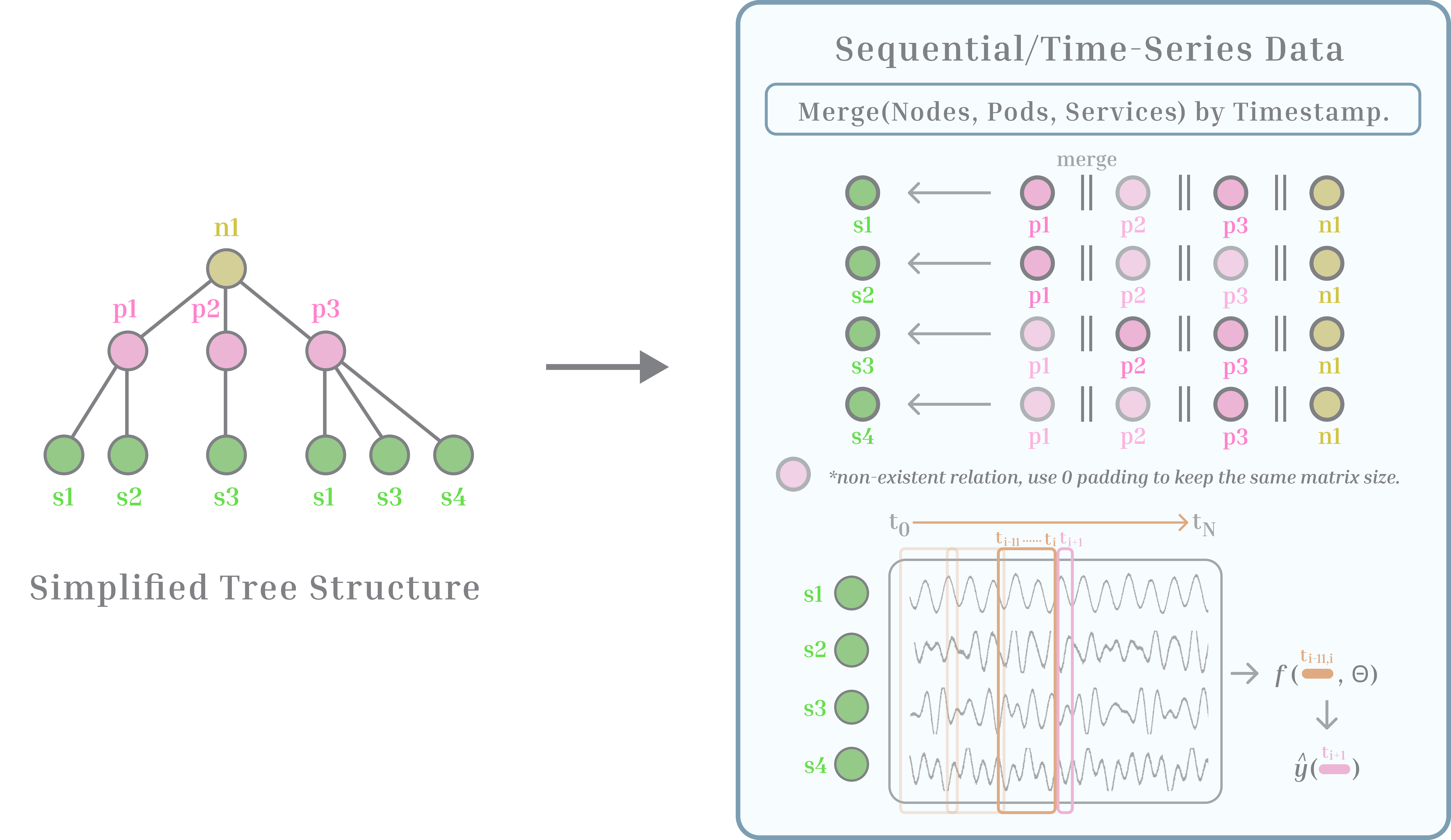}
    \caption{This figure presents the sequential data representation based on the simplified tree structure.}
    \label{fig:D1}
\end{figure}

Operationally, time-evolving attributes can be extracted by querying the nodes and relations from the triples in TTL files. We can then simply merge these data by timestamps to form the sequential data representation. This operation largely ignores the structural relations between nodes, e.g., (Pod, hasNode, Node). 

As shown in the "Sequential Data Representation" part of Fig. \ref{fig:D1}, temporal data from Node 1 is systematically merged into Pods 1, 2, and 3 based on their respective timestamps. Then the merged data from these Pods is further consolidated into Services 1, 2, 3, and 4, ensuring temporal consistency throughout the hierarchical aggregation process.

Mathematically, for each given vertex category $k$, the time-evolving attributes can be represented as a tensor $ \mathbf{\Tilde{X}}_k \in \mathbb{R}^{T \times D_k} $, where $T$ indicates the number of snapshots, and $D_k$ indicates the number of features of category $k$ by concatenating all the instances, which equals to $N_k \times d_k$ with $N_k$ indicates the number of instances of the category $k$, and $d_k$ is the number of features of the category $k$. Then we can merge all vertices Node, Pod and Service together based on timestamps, thus we can get a $ \mathbf{\Tilde{X}} \in \mathbb{R}^{T \times D} $, where $D = D_{Node} + D_{Pod} + D_{Service} = \sum^3_{k=1} N_k* d_k$. 
A sliding window then captures temporal dependencies by concatenating the historical information of the past H time steps, as shown in Eq. \ref{eq:f_vector}.

\begin{equation} \label{eq:f_vector}
    \mathbf{f}(t_j) = [\mathbf{\Tilde{x}}(t_j - H + 1) \, | \, \mathbf{\Tilde{x}}(t_j - H + 2) \, | \, \cdots \, | \, \mathbf{\Tilde{x}}(t_j)]
\end{equation} 

Therefore, the sequential data representation the DKG can be expressed by a matrix $\mathbf{X} \in \mathbb{R}^{T \times D \times H} $, where $ H $ is the number of historical steps, as shown in Eq. \ref{eq:f_matrix}, where $\mathbf{f}(t_j)$ is the aggregated attribute vector at timestamp $t_j$.
 
\begin{equation} \label{eq:f_matrix}
    \mathbf{X} = 
    \begin{bmatrix}
        \mathbf{f}(t_1) \\
        \mathbf{f}(t_2) \\
        \vdots \\
        \mathbf{f}(t_T)
    \end{bmatrix}
    =
    \begin{bmatrix}
        \mathbf{\Tilde{x}}(t_1 - H + 1) & \mathbf{\Tilde{x}}(t_1 - H + 2) & \cdots & \mathbf{\Tilde{x}}(t_1) \\
        \mathbf{\Tilde{x}}(t_2 - H + 1) & \mathbf{\Tilde{x}}(t_2 - H + 2) & \cdots & \mathbf{\Tilde{x}}(t_2) \\
        \vdots & \vdots & \ddots & \vdots \\
        \mathbf{\Tilde{x}}(t_m - H + 1) & \mathbf{\Tilde{x}}(t_m - H + 2) & \cdots & \mathbf{\Tilde{x}}(t_m)
    \end{bmatrix}
\end{equation} 

Once the matrix representation of the attributes $\mathbf{X}$ is obtained, additional time-dependent features can be computed to enhance the model's ability to detect anomalies. For example, calculating the difference between consecutive data points can capture the trend of the time series and understand the direction of change. Similarly calculating the variance of data within a specific time window can reflect the stability of the data. By additionally incorporating the difference and variance as temporal features, the dynamic behaviour of the system can be effectively captured and the ability to identify anomalies can be improved.



Based on this data representation, machine learning algorithms such as SVM, XGB, and IF, as well as deep learning models like MLP, TCN, LSTM, GRU and SA, can use this input to capture information across multiple time steps.

\subsection{Hierarchical Data Representation}

\begin{figure}[!h]
    \centering
    \includegraphics[width=0.65\textwidth]{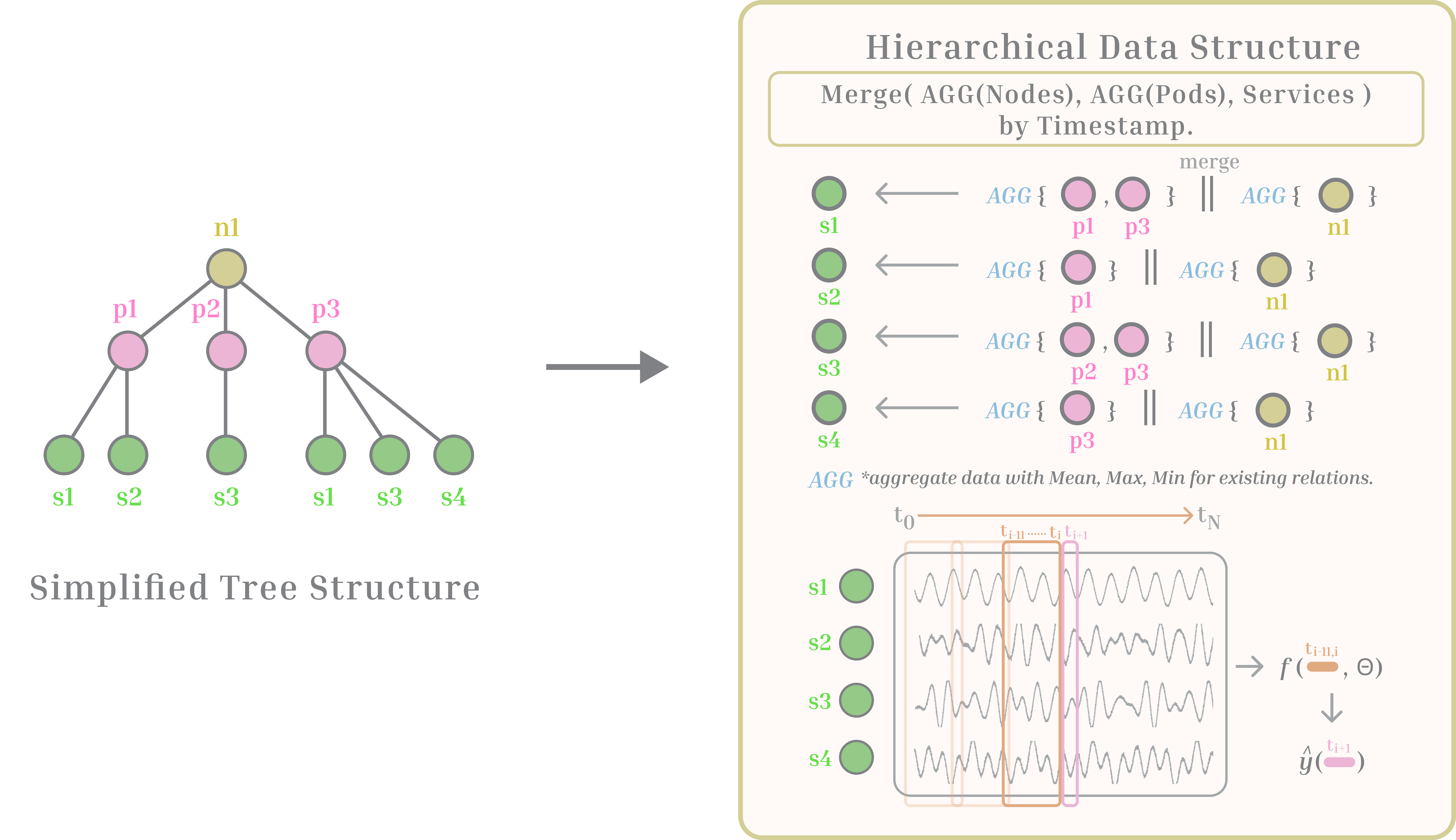}
    \caption{This figure presents the hierarchical data representation based on the simplified tree structure.}
    \label{fig:D2}
\end{figure}

To investigate the structural information within the DKGs, we propose a hierarchical neighbourhood information aggregation method designed to capture the connectivity relationships among various entities. In the simplified tree structure, hierarchical relationships exist between vertices. For example, a Node has multiple Pods and a Pod is connected to different Services through multiple Connections and the "hasConnection" relationship (see Fig. \ref{fig:input}).



Operationally, when relationships exist across Nodes, Pods, and Services, we can progressively aggregate their instances' time-evolving attributes from Nodes to Services using aggregation methods such as mean, max, and min. This approach enables the construction of a structural-temporal representation for each Service instance at each timestamp, as illustrated in the "Hierarchical Data Structure" in Fig. \ref{fig:D2}.

This operation influences the feature dimension $D$ of $\mathbf{X}$, defined in Section \ref{subsec:D1}. Initially, $D$ was determined by $N_k$, the number of instances of category $k$. However, we have now redefined $D$ into $\sum^3_{k=1} AGG * d_k$, with $AGG$ corresponding to the number of aggregation methods employed.

This aggregation process is analogous to a multi-hop operation, where information is aggregated from multiple hops away from the target Service. For instance, for Service 1 Fig. \ref{fig:D2}, temporal data is first aggregated from Node 1, followed by the aggregation of data from Pods 1, 2, and 3, and finally merging them based on the timestamp. This method ensures that the temporal dynamics and structural relationships inherent in the DKG are effectively captured, thereby enhancing the accuracy and robustness of anomaly detection within the MSA.


\subsection{Hierarchical and Inter-Service Representation}

Some graph machine learning algorithms have pointed out that a larger convolutional field can optimise the model \cite{yang2024dynamicsurvey}. Since we have already aggregated the multi-hop structural-temporal information into the Service vertex, in this section we supposed that all Service instances can influence each other. Therefore, we can further enrich the DKG representations by providing additional correlations between Service instances. 

\begin{figure}[!h]
    \centering
    \includegraphics[width=0.65\textwidth]{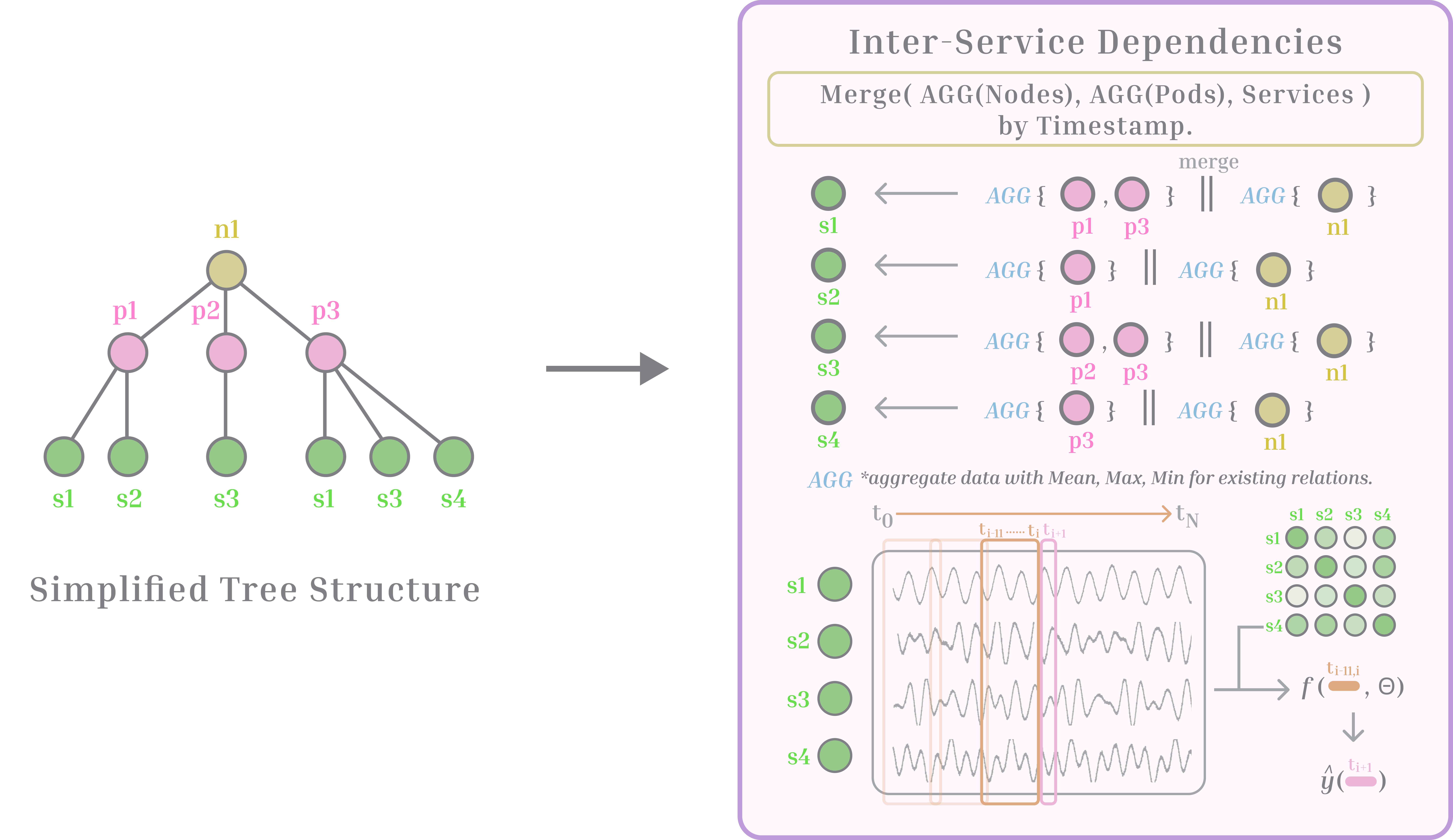}
    \caption{This figure presents the inter-service data representation based on the simplified tree structure.}
    \label{fig:D3}
\end{figure}

This representation can be intuitively achieved by concatenating vertex features, as shown in Eq. \ref{eq:twohop} where $N_\mathcal{S}$ represents the number of Service instances and $\mathbf{s}_i(t)$ represents the data for the $i$-th Service instance at timestamp $t$. However, this approach will greatly increase the size of the feature dimension $D$ of $\mathbf{X}$, requiring more computational resources during feature processing and model training.

\begin{equation} \label{eq:twohop}
    \mathbf{X}_{\text{service}}(t) = [\mathbf{s}_1(t), \mathbf{s}_2(t), \ldots, \mathbf{s_{\mathit{N_S}}}(t)];
\end{equation}

Another method is adding implicitly the inter-service dependencies using the model layers, like Self-Attention, LSTM to capture the correlations among vertices during the model training. SA works by computing the relevance (or attention) between each pair of elements in a sequence, enabling the model to weigh the importance of each element relative to others. This allows the model to focus on relevant parts of the input sequence when making predictions, effectively capturing long-range dependencies and complex relationships within the data. See the "Inter-Service Dependencies" part in Fig. \ref{fig:D3}.

\subsection{Ensemble Learning}

Ensemble learning is a prevalent methodology in machine learning that reduce variance and enhance model robustness by combining multiple models. The effectiveness of ensemble learning depends on the integration of diverse, high-performing models \cite{dietterich2000ensemble}. It allows multiple models to make predictions on the same problem and vote on the final predictions. Voting can be hard or soft voting, with mechanisms such as unanimous and majority voting, which will be introduced in this subsection.

\begin{figure}[!h]
    \centering
    \includegraphics[width=0.75\textwidth]{ensemble2.png}
    \caption{The ensemble learning framework used in this competition. The best-performing models: XGB, SVM, and SA are combined using three voting mechanisms: hard-unanimous, hard-majority, and soft voting, to generate the prediction $\hat{y_1}$, where $\hat{y_1} \in \{0, 1\}$, with 0 indicating normal and 1 indicating anomaly. Optionally, a two-stage model combination can be applied, where the unsupervised Isolation Forest further refines the anomalies detected in $\hat{y_1}$ to reduce false positives.}

    \label{fig:dataset}
\end{figure}

In hard voting, each classifier votes for a class label, and the final prediction is determined by the majority of these classifiers. In soft voting, each classifier provides a probability for each class, and the final prediction is based on the highest average probability across all classifiers.

As two common voting mechanisms combined with hard voting, unanimous voting classifies an instance as an anomaly only if all the models agree. Majority voting, on the other hand, classifies an instance as an anomaly if the majority of models agree.

Intuitively, unanimous voting is more conservative and reduces false positives for anomaly detection, but may also miss some true anomalies. The majority voting method, in contrast, has some flexibility and can improve the detection of true anomalies, but it also preserves the false alarms. Both methods are tested on multiple sets of models and it are discussed in the next section.

In the pipeline, we also explore an optional two-stage model to improve anomaly detection accuracy. In this approach, we first identify potential anomalies by merging the results of the three models through hard or soft voting, and then use these initial predicted anomalies as a new dataset $\mathbf{X}_2$, and subsequently feed $\mathbf{X}_2$ into the unsupervised algorithm Isolation Forest as a secondary filtering mechanism.

The aim of incorporating IF into this two-stage pipeline is to reduce the number of false positives through tree-based segmentation and isolation of outliers. This approach allows us to identify unique, isolated anomalies that may have been inaccurately classified in the first stage, thereby improving the robustness of the overall anomaly detection process.

\section{Experiment} \label{sec:experiment}

\subsection{Experimental Setup}

All experiments were conducted on the ISWC 2024 ADDKG dataset, containing five days of labelled data. Three days were used for training and two for testing, with data sampled at 15 second intervals. The dataset has a significant imbalance, with anomalies representing only 4\% of the total. To tackle this, class weights were incorporated into the loss functions to improve the models' ability of anomaly detection.


As outlined in the methodology section, we tested three levels of representation. 
For the first representation, sequential modelling is used, where the training set is randomly divided into 80\% for training and 20\% for validation. For the other two representations, a more refined data splitting strategy is used. While the 80-20 split is still used for the normal class, anomalies are treated as continuous anomalous events to preserve their integrity. This ensures that the models learn from complete sequences of anomalies. For each anomaly type, one complete event is randomly selected for validation, and the remaining events are used for training.






In the paper, we focus on binary anomaly detection, aiming to distinguish between normal and anomalous data without delving into the specific types of anomalies. The predictions are evaluated by F1 score, precision, and recall.

\subsection{Performance of Models}

This section presents the experimental results for the three data representations:  D1, D2, and D3, correspond respectively to sequential data, hierarchical data, and inter-service data. Table \ref{tab1} summarises the performance of individual models, while table \ref{tab2} focuses on the results of different ensemble methods.


The first part of the experiments evaluates the performance of each model under different data representations, as shown in Table \ref{tab1}. Table \ref{tab1} only showcases the best results for each case. It is worth remarking that the baseline score provided by the challenge organisers is the score obtained by predicting all snapshots given at each timestamp as anomalies, and thus it has a recall equal to 1. The results indicate that incorporating graph structures and service interactions generally enhances model performance, particularly in recall and F1 scores.

While our study aimed to evaluate multiple models comprehensively across different configurations, certain constraints affected the depth of testing. Due to daily submission limits, some models could not be exhaustively tested across all configurations. For instance, XGB and SVM were only evaluated under configuration D2. D3 representation needs to incorporate additional inter-service information, which requires significant computational resources, limiting our ability to test them further. Moreover, the TCN model detected no anomalies following training under D3; as a result, its findings are omitted from this analysis.

\begin{table}[h!]
\caption{Performance of each model across different data representations. The best performance is highlighted in bold.}\label{tab1}
\centering
\scriptsize
\begin{threeparttable}
\begin{tabular}{|l|l|l|l|l|}
\hline
Model &  Dataset Phase & F1 Score & Precision & Recall \\ \hline
Baseline &  - & 0.06759 & 0.03498 & 1.00000 \\ \hline
MLP &  D1 & 0.09390 & 0.05102 & 0.58824\\ \hline
XGB &  D2 & 0.21622 & 0.20000 & 0.23529 \\ \hline
SVM &  D2 & 0.16185 & 0.08974 & 0.82353 \\ \hline
\multirow{2}{*}{TCN} & D1 & 0.09816 & 0.05479 & 0.47059 \\ 
                               & D2 & 0.07792 &  0.04206 & 0.52941 \\ \hline
\multirow{3}{*}{LSTM} & D1 & 0.06759 & 0.03498 & 1.00000 \\ 
                               & D2 & 0.10046 & 0.05446 & 0.64706 \\ 
                               & D3 & 0.16923 & 0.09735 & 0.64706 \\ \hline
\multirow{3}{*}{GRU} & D1 & 0.06759 & 0.03498 & 1.00000 \\ 
                               & D2 & 0.07362 &  0.04110 & 0.35294 \\ 
                               & D3 & 0.16250 & 0.09091 & 0.76471 \\ \hline
\multirow{3}{*}{{\textbf{Self-Attention}}} & D1 & 0.07203 & 0.03736 & 1.00000 \\ 
                                & D2 & 0.09028 & 0.04797 & 0.76471 \\ 
                               &  \textbf{D3} & \textbf{0.22018} & 0.13043 & 0.70588 \\ \hline
\end{tabular}

\footnotesize
\end{threeparttable}
\end{table}

The second part of the experiments evaluates the performance of different ensemble methods, as shown in table \ref{tab2}. Table \ref{tab2} only shows the best results for each case. We tested combinations of XGB, SVM, SA, and IF, along with different voting mechanisms: soft, hard-unanimous, and hard-majority. The results show that the combination of SVM, XGB and SA achieves the best performance in hard-unanimous voting.

\begin{table}
\caption{Performance of different ensemble methods.The best performance is highlighted in bold.}\label{tab2}
\centering
\scriptsize
\begin{threeparttable}
\begin{tabular}{|l|l|l|l|l|l|}
\hline
Ensemble Model & Method & Dataset Phase & F1 Score & Precision & Recall\\ \hline
Baseline &  - &  - & 0.06759 & 0.03498 & 1.00000 \\ \hline
XGB + IF &  soft &  D2 & 0.34286 & 0.33333 & 0.35294 \\ \hline
SVM + IF &  hard,unanimous &  D2 & 0.19802 & 0.11905 & 0.58824 \\ \hline
XGB + SVM + IF &  soft &  D2 & 0.38596 & 0.275 & 0.64706 \\ \hline
XGB + SVM + SA + IF &  hard,unanimous &  D2+D3 & 0.36364 & 0.2963 & 0.47059 \\ \hline
\textbf{XGB + SVM + SA}  &  hard,unanimous &  D2+D3 & {\bfseries0.51429} & 0.5 & 0.52941 \\ \hline

\end{tabular}

\scriptsize
\end{threeparttable}
\end{table}

\begin{figure}[!h]
    \centering
    \includegraphics[width=0.8\textwidth]{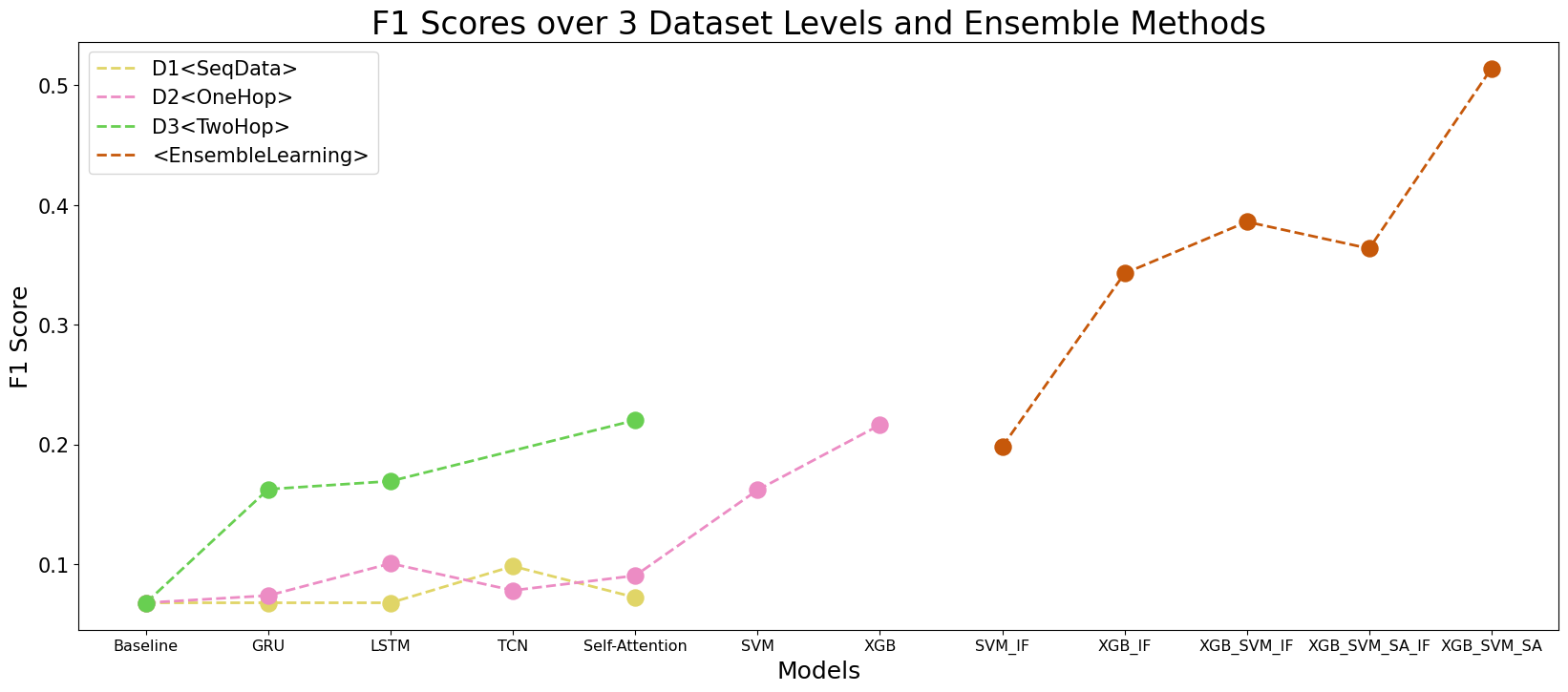}
    \caption{The figure shows the F1 scores of the models across the three data representations in the left, as well as the F1 scores obtained after applying ensemble learning methods in the right. The individual models include XGB, SVM, TCN, LSTM, GRU, and SA Model. The models used in the ensemble learning approach are XGB, IF, SVM, and SA Model.}
    \label{fig:metric_viz}
\end{figure}

Fig. \ref{fig:metric_viz} shows that the hierarchical data representation, inter-service dependency representation as well as ensemble learning methods all significantly outperforms the baseline result, demonstrating the potential of our approach.

\section{Conclusion} \label{sec:conclusion}

In this paper, we explored the effectiveness of various machine learning and deep learning models for anomaly detection on dynamic knowledge graphs, focusing on a Micro-services environment. Using a structured experimental approach, we analysed the performance of the models on three representations: sequential data, hierarchical data, and inter-service dependency data.

Our results show that incorporating graph structures significantly enhances model performance. The inter-service structure, which includes service interactions and hierarchical representation, delivered the best results across most metrics, emphasising the importance of capturing complex structures and (inter-)dependencies. Deep learning models, especially the SA model, performed particularly well in leveraging these structures.

Furthermore, the ensemble methods significantly improved performance by combining the strengths of the top individual models. Unanimous voting, which requires all models to agree on an anomaly, outperformed majority voting in terms of precision, matching the competition's emphasis on precision. 

Future work should focus on several improvements. First, improving model scalability to handle larger, more complex datasets is essential. Second, exploring advanced ensemble techniques such as stacking could improve detection accuracy. Third, integrating real-time anomaly detection would benefit practical applications in dynamic environments. Additionally, our study was limited to binary classification; future research should improve the accuracy of detecting different types of anomalies. Finally, although we used graph structure information, the application of graph neural networks remains a promising direction for further investigation. Moreover, our data representations are high-dimensional, we can compare and use some low-dimensional representations, such as node embedding, graph embedding to test our proposed solutions.

This study highlights the potential of combining graph-based data representations with advanced machine learning and deep learning techniques for anomaly detection on dynamic knowledge graphs. Refining these methods and exploring the suggested future directions can lead to more robust, accurate, and scalable solutions for complex anomaly detection in various domains.




\begin{thebibliography}{00}
\bibitem{bronstein2017geometric}Bronstein, M., Bruna, J., LeCun, Y., Szlam, A. \& Vandergheynst, P. Geometric deep learning: going beyond euclidean data. {\em IEEE Signal Processing Magazine}. \textbf{34}, 18-42 (2017)
\bibitem{yu2017spatio}Yu, B., Yin, H. \& Zhu, Z. Spatio-temporal graph convolutional networks: A deep learning framework for traffic forecasting. {\em ArXiv Preprint ArXiv:1709.04875}. (2017)
\bibitem{gilmer2017neural}Gilmer, J., Schoenholz, S., Riley, P., Vinyals, O. \& Dahl, G. Neural message passing for quantum chemistry. {\em International Conference On Machine Learning}. pp. 1263-1272 (2017)
\bibitem{shi2022heterogeneous}Shi, C. Heterogeneous Graph Neural Networks. {\em Graph Neural Networks: Foundations, Frontiers, And Applications}. pp. 351-369 (2022)
\bibitem{yang2024dynamicsurvey}Yang, L., Chatelain, C. \& Adam, S. Dynamic Graph Representation Learning With Neural Networks: A Survey. {\em IEEE Access}. \textbf{12} pp. 43460-43484 (2024)
\bibitem{SVM}Schölkopf, B., Williamson, R., Smola, A., Shawe-Taylor, J. \& Platt, J. Support vector method for novelty detection. {\em Advances In Neural Information Processing Systems}. \textbf{12} (1999)
\bibitem{IF}Liu, F., Ting, K. \& Zhou, Z. Isolation forest. {\em 2008 Eighth Ieee International Conference On Data Mining}. pp. 413-422 (2008)
\bibitem{XGB}Chen, T. \& Guestrin, C. Xgboost: A scalable tree boosting system. {\em Proceedings Of The 22nd Acm Sigkdd International Conference On Knowledge Discovery And Data Mining}. pp. 785-794 (2016)
\bibitem{RNNJordanNetwork}Servan-Schreiber, D., Cleeremans, A. \& McClelland, J. Graded state machines: The representation of temporal contingencies in simple recurrent networks. {\em Machine Learning}. \textbf{7} pp. 161-193 (1991)
\bibitem{LSTM}Hochreiter, S. \& Schmidhuber, J. Long short-term memory. {\em Neural Computation}. \textbf{9}, 1735-1780 (1997)
\bibitem{GRU}Cho, K., Van Merriënboer, B., Gulcehre, C., Bahdanau, D., Bougares, F., Schwenk, H. \& Bengio, Y. Learning phrase representations using RNN encoder-decoder for statistical machine translation. {\em ArXiv Preprint ArXiv:1406.1078}. (2014)
\bibitem{oord2016wavenet}Oord, A., Dieleman, S., Zen, H., Simonyan, K., Vinyals, O., Graves, A., Kalchbrenner, N., Senior, A. \& Kavukcuoglu, K. Wavenet: A generative model for raw audio. {\em ArXiv Preprint ArXiv:1609.03499}. (2016)
\bibitem{vaswani2017attentionneed}Vaswani, A., Shazeer, N., Parmar, N., Uszkoreit, J., Jones, L., Gomez, A., Kaiser, L. \& Polosukhin, I. Attention Is All You Need. {\em ArXiv Preprint ArXiv:1706.03762}. (2017,6), https://arxiv.org/abs/1706.03762
\bibitem{dietterich2000ensemble}Dietterich, T. Ensemble methods in machine learning. {\em International Workshop On Multiple Classifier Systems}. pp. 1-15 (2000)
\bibitem{SurveyGADma}Ma, X., Wu, J., Xue, S., Yang, J., Zhou, C., Sheng, Q., Xiong, H. \& Akoglu, L. A comprehensive survey on graph anomaly detection with deep learning. {\em IEEE Transactions On Knowledge And Data Engineering}. \textbf{35}, 12012-12038 (2021)
\bibitem{DGADMTS}Chen, K., Feng, M. \& Wirjanto, T. Multivariate Time Series Anomaly Detection via Dynamic Graph Forecasting. {\em ArXiv Preprint ArXiv:2302.02051}. (2023)
\bibitem{StrGNN}Cai, L., Chen, Z., Luo, C., Gui, J., Ni, J., Li, D. \& Chen, H. Structural temporal graph neural networks for anomaly detection in dynamic graphs. {\em Proceedings Of The 30th ACM International Conference On Information \& Knowledge Management}. pp. 3747-3756 (2021)
\bibitem{AddGraph}Zheng, L., Li, Z., Li, J., Li, Z. \& Gao, J. AddGraph: Anomaly Detection in Dynamic Graph Using Attention-based Temporal GCN.. {\em IJCAI}. \textbf{3} pp. 7 (2019)
\bibitem{TADDY}Liu, Y., Pan, S., Wang, Y., Xiong, F., Wang, L., Chen, Q. \& Lee, V. Anomaly detection in dynamic graphs via transformer. {\em IEEE Transactions On Knowledge And Data Engineering}. (2021)
\bibitem{IOTsteenwinckel2022data}Steenwinckel, B., De Brouwer, M., Stojchevska, M., Van Der Donckt, J., Nelis, J., Ruyssinck, J., Herten, J., Casier, K., Van Ooteghem, J., Crombez, P. \& Others Data analytics for health and connected care: Ontology, knowledge graph and applications. {\em International Conference On Pervasive Computing Technologies For Healthcare}. pp. 344-360 (2022)
\bibitem{IOTsteenwinckel2021flags}Steenwinckel, B., De Paepe, D., Hautte, S., Heyvaert, P., Bentefrit, M., Moens, P., Dimou, A., Van Den Bossche, B., De Turck, F., Van Hoecke, S. \& Others FLAGS: A methodology for adaptive anomaly detection and root cause analysis on sensor data streams by fusing expert knowledge with machine learning. {\em Future Generation Computer Systems}. \textbf{116} pp. 30-48 (2021)
\bibitem{dorpinghaus2022social}Dörpinghaus, J., Klante, S., Christian, M., Meigen, C. \& Düing, C. From social networks to knowledge graphs: A plea for interdisciplinary approaches. {\em Social Sciences \& Humanities Open}. \textbf{6}, 100337 (2022)
\bibitem{TGN}Rossi, E., Chamberlain, B., Frasca, F., Eynard, D., Monti, F. \& Bronstein, M. Temporal graph networks for deep learning on dynamic graphs. {\em ArXiv Preprint ArXiv:2006.10637}. (2020)
\bibitem{bca}Kumar, S., Spezzano, F., Subrahmanian, V. \& Faloutsos, C. Edge weight prediction in weighted signed networks. {\em Data Mining (ICDM), 2016 IEEE 16th International Conference On}. pp. 221-230 (2016)
\bibitem{bco}Kumar, S., Hooi, B., Makhija, D., Kumar, M., Faloutsos, C. \& Subrahmanian, V. Rev2: Fraudulent user prediction in rating platforms. {\em Proceedings Of The Eleventh ACM International Conference On Web Search And Data Mining}. pp. 333-341 (2018)
\bibitem{hodge2004survey}Hodge, V. \& Austin, J. A survey of outlier detection methodologies. {\em Artificial Intelligence Review}. \textbf{22} pp. 85-126 (2004)

\bibitem{adgnn_survey}Kim, H., Lee, B., Shin, W. \& Lim, S. Graph Anomaly Detection With Graph Neural Networks: Current Status and Challenges. {\em IEEE Access}. \textbf{10} pp. 111820-111829 (2022)

\bibitem{dominant}Ding, K., Li, J., Bhanushali, R. \& Liu, H. Deep Anomaly Detection on Attributed Networks. {\em Proceedings Of The 2019 SIAM International Conference On Data Mining (SDM)}. pp. 594-602, https://epubs.siam.org/doi/abs/10.1137/1.9781611975673.67

\bibitem{cola}Liu, Y., Li, Z., Pan, S., Gong, C., Zhou, C. \& Karypis, G. Anomaly Detection on Attributed Networks via Contrastive Self-Supervised Learning. {\em IEEE Transactions On Neural Networks And Learning Systems}. \textbf{33}, 2378-2392 (2022,6), http://dx.doi.org/10.1109/TNNLS.2021.3068344

\bibitem{glocal}Ma, R., Pang, G., Chen, L. \& Hengel, A. Deep Graph-level Anomaly Detection by Glocal Knowledge Distillation. {\em Proceedings Of The Fifteenth ACM International Conference On Web Search And Data Mining}. pp. 704-714 (2022,2), http://dx.doi.org/10.1145/3488560.3498473

\bibitem{ocgtl}Qiu, C., Kloft, M., Mandt, S. \& Rudolph, M. Raising the Bar in Graph-level Anomaly Detection.  (2022), https://arxiv.org/abs/2205.13845

\bibitem{newman2021building}Newman, S. Building microservices. (" O'Reilly Media, Inc.",2021)

\bibitem{yagoub2018equipment}Yagoub, I., Khan, M. \& Jiyun, L. IT equipment monitoring and analyzing system for forecasting and detecting anomalies in log files utilizing machine learning techniques. {\em 2018 International Conference On Advances In Big Data, Computing And Data Communication Systems (icABCD)}. pp. 1-6 (2018)

\bibitem{sharma2013cloudpd}Sharma, B., Jayachandran, P., Verma, A. \& Das, C. CloudPD: Problem determination and diagnosis in shared dynamic clouds. {\em 2013 43rd Annual IEEE/IFIP International Conference On Dependable Systems And Networks (DSN)}. pp. 1-12 (2013)


\bibitem{pahl2018all}Pahl, M. \& Aubet, F. All eyes on you: Distributed Multi-Dimensional IoT microservice anomaly detection. {\em 2018 14th International Conference On Network And Service Management (CNSM)}. pp. 72-80 (2018)

\bibitem{knn}Peterson, L. K-nearest neighbor. {\em Scholarpedia}. \textbf{4}, 1883 (2009)
\bibitem{rf}Ho, T. Random decision forests. {\em Proceedings Of 3rd International Conference On Document Analysis And Recognition}. \textbf{1} pp. 278-282 (1995)
\bibitem{mlp}Haykin, S. Neural networks: a comprehensive foundation. (Prentice Hall PTR,1994)

\bibitem{dkg}Alam, M., Gesese, G. \& Paris, P. Neurosymbolic Methods for Dynamic Knowledge Graphs.  (2024), https://arxiv.org/abs/2409.04572
\bibitem{kg}Ji, S., Pan, S., Cambria, E., Marttinen, P. \& Yu, P. A Survey on Knowledge Graphs: Representation, Acquisition, and Applications. {\em IEEE Transactions On Neural Networks And Learning Systems}. \textbf{33}, 494-514 (2022)

\end{thebibliography}

\end{document}